\pdfoutput=1

\PassOptionsToPackage{table,xcdraw}{xcolor}

\documentclass[11pt]{article}

\usepackage{EMNLP2022}

\usepackage{times}
\usepackage{latexsym}

\usepackage[T1]{fontenc}

\usepackage[utf8]{inputenc}

\usepackage{microtype}

\usepackage{inconsolata}

\usepackage{graphicx}
\usepackage{booktabs}
\usepackage{multirow}
\usepackage[normalem]{ulem}
\useunder{\uline}{\ul}{}
\usepackage{xurl}
\usepackage[page,header]{appendix}
\usepackage{caption}
\usepackage{subcaption}
\usepackage{amsmath}

\title{IDK-MRC: Unanswerable Questions for Indonesian Machine Reading Comprehension}

\author{Rifki Afina Putri \\
  School of Computing \\
  KAIST, South Korea \\
  \texttt{rifkiaputri@kaist.ac.kr} \\\And
  Alice Oh \\
  School of Computing \\
  KAIST, South Korea \\
  \texttt{alice.oh@kaist.edu} \\}

\begin{document}
\maketitle
\begin{abstract}

Machine Reading Comprehension (MRC) has become one of the essential tasks in Natural Language Understanding (NLU) as it is often included in several NLU benchmarks \cite{liang-etal-2020-xglue,wilie-etal-2020-indonlu}. However, most MRC datasets only have answerable question type, overlooking the importance of unanswerable questions. MRC models trained only on answerable questions will select the span that is most likely to be the answer, even when the answer does not actually exist in the given passage \cite{rajpurkar-etal-2018-know}. This problem especially remains in medium- to low-resource languages like Indonesian. Existing Indonesian MRC datasets \cite{purwarianti-etal-2007-machine,clark-etal-2020-tydi} are still inadequate because of the small size and limited question types, i.e., they only cover answerable questions. To fill this gap, we build a new Indonesian MRC dataset called I(n)don'tKnow-MRC (IDK-MRC) by combining the automatic and manual unanswerable question generation to minimize the cost of manual dataset construction while maintaining the dataset quality. Combined with the existing answerable questions, IDK-MRC consists of more than 10K questions in total. Our analysis shows that our dataset significantly improves the performance of Indonesian MRC models, showing a large improvement for unanswerable questions\footnote{The code and dataset of IDK-MRC are available at \url{https://github.com/rifkiaputri/IDK-MRC}}.
\end{abstract}

\section{Introduction}
Machine Reading Comprehension (MRC) is a task where a machine is asked to read a given passage and answer a question based on the passage. Several English MRC datasets have been widely used, including SQuAD \cite{rajpurkar-etal-2016-squad} and NewsQA \cite{trischler-etal-2017-newsqa}. However, MRC models that do well on those datasets are not guaranteed to be robust. \citet{rajpurkar-etal-2018-know} highlights the problem of the SQuAD dataset that only focuses on answerable questions, making the model trained on this dataset tends to select the span without carefully checking whether the passage actually has the answer. SQuAD 2.0 is then built by manually adding new unanswerable questions to the existing SQuAD dataset \cite{rajpurkar-etal-2018-know}.

\begin{figure}[t!]
    \begin{center}
        \centerline{\includegraphics[width=\linewidth]{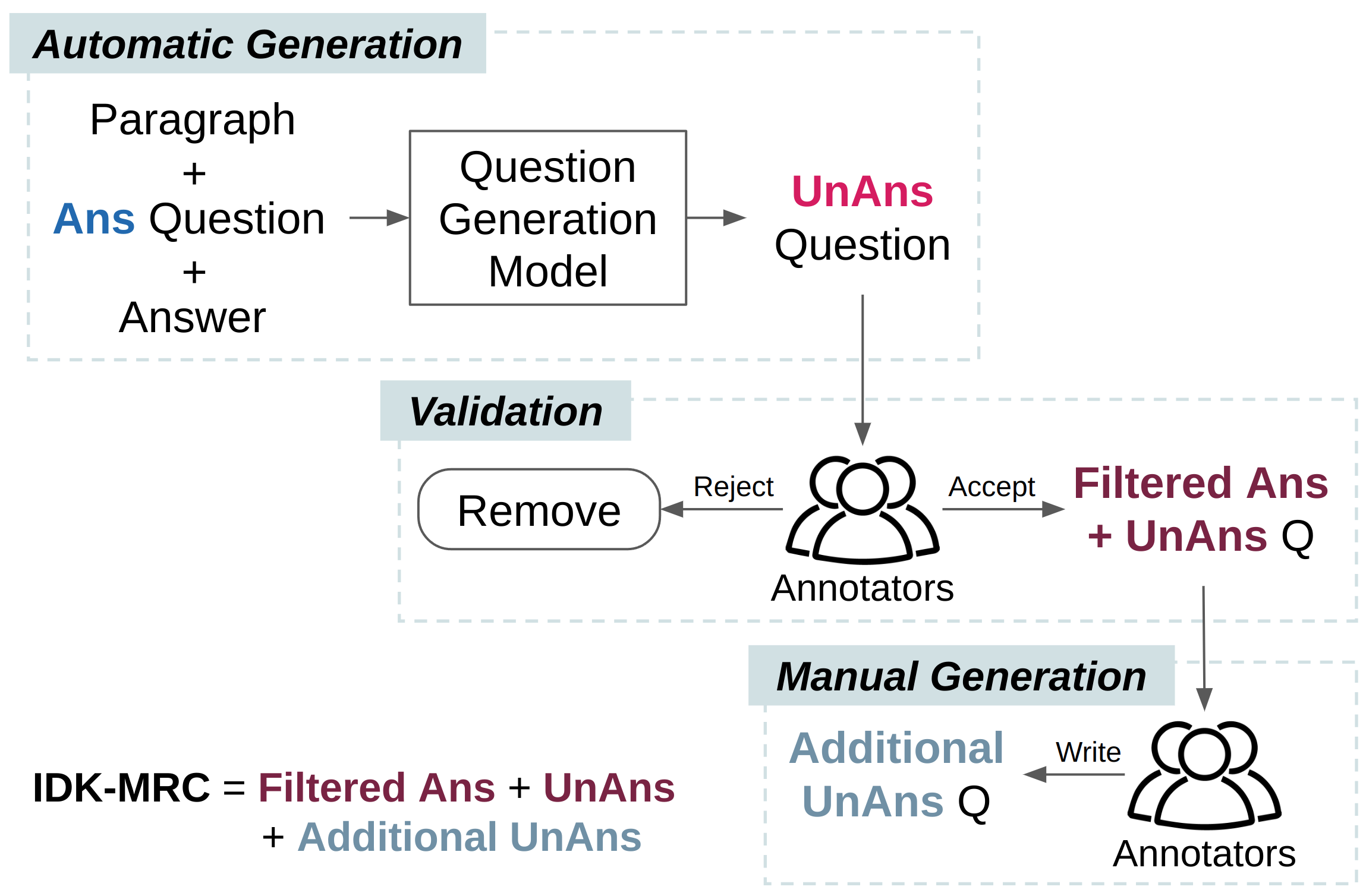}}
        \caption{Our dataset collection pipeline.}
        \label{fig:data_collection}
    \end{center}
\end{figure}

\begin{table*}[t!]
\centering
\begin{small}
\begin{tabular}{@{}p{0.1\textwidth}p{0.15\textwidth}rp{0.58\textwidth}@{}}
\toprule
\textbf{Type} &
  \multicolumn{1}{c}{\textbf{Description}} &
  \multicolumn{2}{c}{\textbf{Example}} \\ \midrule
Negation &
  \multirow{3}{7em}{Negation word inserted or removed} &
  Context &
  Kambing memiliki lemak dalam kandungan susunya. \textit{(Goats have fat in their milk.)} \\
 &
   &
  Ans Q &
  Apakah kandungan yang ada dalam susu kambing? \textit{(What are the ingredients in goat's milk?)} \\
 &
   &
  UnAns Q &
  Apakah kandungan yang \textbf{tidak} ada dalam susu kambing? \textit{(What are the ingredients that \textbf{do not} exist in goat's milk?)} \\ \midrule
Antonym &
  Antonym used &
  Context &
  Aristokrasi adalah sebuah kelas sosial yang \textbf{tertinggi} di masyarakat. \textit{(Aristocracy is the \textbf{highest} social class in society.)} \\
 &
   &
  Ans Q &
  Apakah nama kelas sosial \textbf{tertinggi}? \textit{(What is the name of the \textbf{highest} social class?)} \\
 &
   &
  UnAns Q &
  Apa nama kelas sosial \textbf{terendah}? \textit{(What is the name of the \textbf{lowest} social class?)} \\ \midrule
Entity Swap &
  \multirow{3}{7em}{Entity, date, number, or term replaced with other entity, date, number, or term} &
  Context &
  Salah satu kandidat standar untuk 4G yang dikomersilkan di dunia yaitu standar Long Term Evolution (LTE) (Swedia sejak 2009). \textit{(One of the standards for 4G commercialized in the world is the Long Term Evolution (LTE) standard (Sweden since 2009).)} \\
 &
   &
  Ans Q &
  Di manakah \textbf{LTE} pertama kali diciptakan? \textit{(Where was \textbf{LTE} first invented?)} \\
 &
   &
  UnAns Q &
  Di manakah \textbf{3G} pertama diciptakan? \textit{(Where was \textbf{3G} first invented?)} \\ \midrule
Question Tag Swap &
  Question tag replaced with other question tag &
  Context &
  Suaka margasatwa Muara Angke adalah sebuah kawasan konservasi di wilayah hutan bakau di pesisir utara Jakarta. \textit{(Muara Angke Wildlife Sanctuary is a conservation area in the mangrove forest area on the north coast of Jakarta.)} \\
 &
   &
  Ans Q &
  \textbf{Di mana} Suaka margasatwa Muara Angke dibangun? \textit{(\textbf{Where} was Muara Angke Wildlife Sanctuary built?)} \\
 &
   &
  UnAns Q &
  \textbf{Kapan} Suaka margasatwa Muara Angke dibangun? \textit{(\textbf{When} was Muara Angke Wildlife Sanctuary built?)} \\ \midrule
Specific Condition &
  \multirow{3}{7em}{Asks for specific condition that is not satisfied by the information in the paragraph} &
  Context &
  Bon Jovi terdiri dari Vokalis Jon Bon Jovi, Keyboardist David Bryan, Drummer Tico Torres, Gitaris Phil X, dan Bassist Hugh McDonald. \textit{(Bon Jovi consists of Vocalist Jon Bon Jovi, Keyboardist David Bryan, Drummer Tico Torres, Guitarist Phil X, and Bassist Hugh McDonald.)} \\
 &
   &
  Ans Q &
  Siapa personil Bon Jovi? \textit{(Who are the members of Bon Jovi?)} \\
 &
   &
  UnAns Q &
  Siapa personil Bon Jovi \textbf{yang paling jarang dikenal}? \textit{(Who is \textbf{the least known member} of Bon Jovi?)} \\ \midrule
Other &
  \multirow{3}{7em}{Other cases where the paragraph does not imply any answer} &
  Context &
  Patrick Star adalah seekor bintang laut yang bersahabat dengan Spongebob. \textit{(Patrick Star is a starfish whose best friend is Spongebob.)} \\
 &
   &
  Ans Q &
  Siapakah \textbf{teman baik} karakter SpongeBob SquarePants? \textit{(Who is SpongeBob SquarePants' \textbf{best friend}?)} \\
 &
   &
  UnAns Q &
  Siapa \textbf{teman kecil} karakter Spongebob SquarePants? \textit{(Who is Spongebob SquarePants' \textbf{childhood friend}?)} \\ \bottomrule
\end{tabular}
\end{small}
\caption{Unanswerable question types that are covered in IDK-MRC.}
\label{tab:data-sample}
\end{table*}

While SQuAD 2.0 is widely used for evaluation of English models, similar datasets for other languages are still limited, hindering the progress of MRC task for these languages. Indonesian has around 198 million speakers\footnote{\url{https://www.babbel.com/en/magazine/how-many-people-speak-indonesian-where-is-it-spoken} (Accessed Jan 2022)}, but despite its popularity, there exists an insufficient amount of Indonesian MRC datasets. For instance, FacQA dataset \cite{purwarianti-etal-2007-machine} has only around 3K samples, and TyDiQA-GoldP dataset \cite{clark-etal-2020-tydi} has around 5K samples. Furthermore, both datasets only have answerable question type, ignoring the importance of incorporating unanswerable questions. Therefore, building an Indonesian MRC dataset that covers unanswerable questions is necessary.

One alternative to construct a new dataset is manually adding the unanswerable questions. This, however, is expensive and time-consuming. Several Question Generation (QG) approaches have been proposed to mitigate this, but most are focused on generating answerable questions \cite{heilman-smith-2010-good,du-etal-2017-learning,du-cardie-2018-harvesting,klein2019learning,alberti-etal-2019-synthetic,kumar-etal-2019-cross,puri-etal-2020-training,shakeri-etal-2020-end}, with only one generating unanswerable questions \cite{zhu-etal-2019-learning}. These models can quickly generate many questions, but the resulting questions are usually less fluent and less relevant to the passage than human-written questions.

This work intends to combine the best of both worlds by incorporating humans into the automatic dataset generation pipeline. Figure \ref{fig:data_collection} shows our pipeline, which consists of three phases: automatic generation, validation, and manual generation. To sum up, our contributions are as follows:
\begin{itemize}
    \item We construct a new Indonesian MRC dataset called I(n)don’tKnow-MRC (IDK-MRC), consisting of over 5K unanswerable questions with diverse question types, as shown in Table \ref{tab:data-sample}. To the best of our knowledge, IDK-MRC is the first Indonesian MRC dataset covering answerable and unanswerable questions.
    \item We propose a simple dataset collection pipeline consisting of automatic and manual dataset generation. We show that relying \textit{only} on automatic generation results in highly imbalanced question type distribution; our manual generation method covers this limitation.
    \item We validate our dataset on the downstream task and show that it effectively improves the MRC models' performance, especially in predicting the answer to the unanswerable questions.
\end{itemize}

\section{Related Work}
\paragraph{Existing Indonesian MRC Dataset}
While many MRC datasets are available in English \cite{rajpurkar-etal-2016-squad,trischler-etal-2017-newsqa,rajpurkar-etal-2018-know}, the number of publicly available Indonesian MRC datasets is very limited. A shortcut to obtain Indonesian MRC data is by machine translating English MRC dataset \cite{muis2020seq2seq}, but it will result in translation artifacts. We may avoid this by recruiting human annotators to translate them manually; still, it leads to \textit{translationese}, where the translated text appears awkward or unnatural \cite{clark-etal-2020-tydi}. FacQA \cite{purwarianti-etal-2007-machine} is part of the IndoNLU benchmark \cite{wilie-etal-2020-indonlu} built from a news article. It has around 3K answerable questions, with limited categories of questions: date, location, name, organization, person, and quantitative. Another dataset called TyDiQA-GoldP \cite{clark-etal-2020-tydi}, a multilingual QA dataset constructed from Wikipedia, has about 5K Indonesian samples. It also only focuses on answerable questions. To this date, there are no publicly available Indonesian MRC datasets that include unanswerable question type.

\paragraph{Human-Model Dataset Construction}
Combining human and model in dataset construction is mainly applied to adversarial data, such as AdversarialQA \cite{bartolo-etal-2020-beat} and AdversarialNLI \cite{nie-etal-2020-adversarial}. In this dynamic adversarial data collection, human annotators are asked to construct adversarial questions to fool the model. Such human-model annotation pipeline has not been tried for unanswerable questions. \citet{wang-etal-2021-want-reduce} analyzed the cost of different dataset labeling strategies, including the combination of GPT-3 \cite{NEURIPS2020_1457c0d6} and human labeling. Although they included the MRC task in their experiment, they only focused on SQuAD 1.1 \cite{rajpurkar-etal-2016-squad}, which only has answerable questions. The effectiveness of the human-model labeling in the context of unanswerable questions remains unclear.

\paragraph{Unanswerable Question Generation}
Various approaches have been proposed for generating answerable questions in English \cite{heilman-smith-2010-good,du-etal-2017-learning, du-cardie-2018-harvesting,lewis-etal-2019-unsupervised,klein2019learning, alberti-etal-2019-synthetic,puri-etal-2020-training,shakeri-etal-2020-end}, Indonesian \cite{muis2020seq2seq}, and cross- or multi-lingual \cite{kumar-etal-2019-cross,chi2020cross,shakeri-etal-2021-towards,riabi-etal-2021-synthetic}. The question generation technique also applied to generate adversarial questions \cite{bartolo-etal-2021-improving}. However, for unanswerable question generation, the number of works are limited. \citet{zhu-etal-2019-learning} proposed Pair-to-Sequence (Pair2Seq) model that uses separate encoders for the paragraph and answerable question. They utilized English word embedding (i.e., GloVe \cite{pennington-etal-2014-glove}) and character embedding as the feature and bi-LSTM \cite{hochreiter1997long} as the encoder. Although their model performed better compared to the rule-based and TF-IDF baselines, it still relied on a traditional word embedding representation as the feature.  Differing from their approach, we utilized mT5 model \cite{xue-etal-2021-mt5} that covers contextual representation of 101 languages, including Indonesian. Our experiment (\S\ref{sec:qgen-experiment}) confirms that our model outperforms Pair2Seq model, demonstrating the advantage of our approach.

\section{Dataset Collection Pipeline}
We build IDK-MRC dataset by combining model-generated unanswerable questions with human-written questions. As shown in Figure \ref{fig:data_collection}, our dataset collection has three stages: automatic generation, validation, and manual generation.

\begin{figure}[t!]
    \begin{center}
        \centerline{\includegraphics[width=5cm]{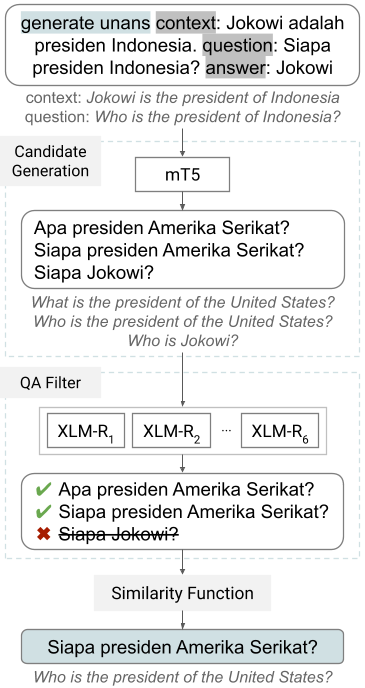}}
        \caption{Our proposed question generation model.}
        \label{fig:qgen_model}
    \end{center}
\end{figure}

\subsection{Automatic Generation}
\label{sec:automatic-gen}

In this stage, we automatically construct unanswerable questions using a Question Generation (QG) model. We use translated SQuAD 2.0 \cite{rajpurkar-etal-2018-know} as the training data of the QG model. In the inference step, we use the answerable questions from TyDiQA-GoldP \cite{clark-etal-2020-tydi} as a starting point to add more unanswerable questions for our dataset. Our QG model architecture is illustrated in Figure \ref{fig:qgen_model}.

\paragraph{Candidate Generation}
We utilize mT5 model \cite{xue-etal-2021-mt5} to generate the unanswerable question candidates. We apply \texttt{generate unans} prefix, followed by \textit{context}, \textit{answerable question}, and \textit{answer} as the input. Then, using top-p and top-k sampling as the decoding method, the model produces several output candidates.

\paragraph{QA Filter}
Since not all output candidates are valid unanswerable questions, we filter out invalid questions using an ensemble of six\footnote{6 was chosen based on related work in Adversarial QA \cite{bartolo-etal-2021-improving}.} Question Answering (QA) models. We fine-tuned XLM-R \cite{conneau-etal-2020-unsupervised} on translated SQuAD 2.0 dataset using different random seeds and used them as the QA models. Based on the prediction of these models, we keep the question if four or more models give an empty answer (i.e., unanswerable) or if four or more models return non-empty answers and all these answers are different.

\paragraph{Similarity Function}
Finally, we apply a similarity function to all remaining output candidates to make sure that the final output is relevant to its corresponding paragraph and answerable question. We calculate similarity between the original answerable question and the remaining question candidates using BLEU score to get the unanswerable question with highest n-gram overlap. We pick the candidate with the highest score as the final output.

\subsection{Validation}
\label{sec:validation}
After obtaining the automatically generated unanswerable questions, we validate them to ensure that they do not have noise or error. We recruit four Indonesian native speakers with 2+ years of experience in Indonesian NLP dataset annotation. Each annotator is asked to give a score to the generated questions with three criteria, adopted from \citet{zhu-etal-2019-learning} and re-defined as follows:
\begin{itemize}
\item \textbf{Unanswerability}: whether the answer can be found in the given paragraph. The score is 1 if the answer cannot be found, 0 otherwise.
\item \textbf{Relevancy}: whether the question is relevant to the paragraph and the answerable question. 3 if the question is relevant to both, 2 if it is only relevant to the paragraph or the answerable question, and 1 if it is not relevant to either. 
\item \textbf{Fluency}: whether the question is fluent. 3 if the collective quality of all words in the question is fluent and coherent; 2 if the question is semi-coherent, has a minor typo, or grammatical errors; and 1 if the question is incoherent or incomprehensible.
\end{itemize}

Each question is validated by one annotator, with each annotator validating the same number of questions. Then, we apply cross-checking method to minimize human errors and to ensure consistency of the criteria across the annotators. Suppose that we have four annotators $(a_1, ..., a_4)$, who have evaluated some set of samples $(s_1, ..., s_4)$. Each sample $s_i$ consists of a set of \textit{paragraph}, \textit{answerable}, \textit{unanswerable question}, along with the \textit{unanswerability}, \textit{relevancy}, and \textit{fluency} scores of the  unanswerable question. In the cross-checking phase, $a_1$ is assigned to check the scores of $s_2$, $a_2$ is assigned to check the scores of $s_1$, and so on. The disagreement\footnote{Overall, the disagreement percentage is roughly around 10--20\%, with $\sim$84\% of the disagreement are categorized as narrow disagreement (1 vs 2 or 2 vs 3).} is resolved by discussion among the annotators to ensure each annotator has the same level of task understanding and thus resulting in high quality and consistent annotation.

Finally, we keep the questions with perfect unanswerability, relevancy, and fluency scores (i.e., questions with scores of 1, 3, 3). We also keep the questions with scores of (1, 3, 2) and ask the annotators to make minor corrections to those questions. We regard the rest of the automatically generated questions as noise and discard them\footnote{We remove irrelevant questions (questions with relevancy score of 1 or 2) because they are often too far from the context, making the task less challenging. For example, the context describes Ecology definition, the Ans Q is \textit{"what is the definition of \textbf{ecology}?"}, and the UnAns Q is \textit{"what is the definition of \textbf{neo}?"}. We remove this kind of question. Note that we still regard the entity as relevant (and keep the question) if it belongs to the same category, i.e., person name, country, etc.}. From 6,196 generated questions, 3,190 questions are kept in the dataset, with 2,840 questions have a perfect score, and 350 questions have 1, 3, and 2 scores for unanswerability, relevancy, and fluency. 

\subsection{Manual Generation}
\label{sec:manual-gen}
In the final stage, we ask human annotators to add more unanswerable questions, especially for the question types that QG model struggles to generate. There are six question types, as listed in Table \ref{tab:data-sample}, and it is important to have a sufficient number of questions for each type. The model generates entity swap questions well (see Figure \ref{fig:test_distribution}), so we request the annotators to write the remaining question types, i.e., \textit{negation}, \textit{antonym}, \textit{question tag swap}, \textit{specific condition}, and \textit{other}. The annotators may also add a new answerable question to be paired with the \textit{negation} question, specifically for the case when the negation word in the answerable question is removed. The annotators are the same as those from the validation stage, and they were assigned to write 500 unanswerable questions each. We also apply the same cross-checking method as the validation stage. In total, we have 2,000 human-written unanswerable questions.

\section{The Resulting Data}
As shown in Table \ref{tab:data-stat}, we have different dataset variations from each dataset collection stage:
\begin{itemize}
    \item \textbf{TyDiQA}: original answerable questions from TyDiQA-GoldP \cite{clark-etal-2020-tydi}\footnote{All questions in TyDiQA-GoldP are extractive.}. Since the test set is not publicly shared, we made a new split from the existing train and dev data.
    \item \textbf{Model Gen}: unanswerable questions output from the automatic generation stage (\S\ref{sec:automatic-gen}) combines with the answerable questions from TyDiQA. We remove questions that are same as the answerable questions\footnote{The questions that are the same as the answerable questions are still exist due to the problem in the decoding process in the automatic generation process. We attempt to eliminate this by adding QA Filter model; however, the QA model is not 100\% perfect. Therefore, this kind of questions can still remain after the automatic filtering.}.
    \item \textbf{Human Filt}: Model Gen dataset that has been manually filtered and validated (\S\ref{sec:validation}).
    \item \textbf{IDK-MRC}: final version of our dataset consisting of Human Filt dataset with additional questions written by annotators in the manual generation stage (\S\ref{sec:manual-gen}).
\end{itemize}

Our final IDK-MRC dataset is the largest publicly available Indonesian MRC dataset with various types of unanswerable questions.

\begin{table}[t!]
    \centering
    \resizebox{0.86\columnwidth}{!}{%
    \begin{tabular}{@{}llr@{\hskip3pt}r@{\hskip3pt}r@{}}
        \toprule
        \multicolumn{1}{c}{\textbf{Split}} & \multicolumn{1}{c}{\textbf{Type}} & \multicolumn{1}{r}{\textbf{Ans}} & \multicolumn{1}{r}{\textbf{UnAns}} & \multicolumn{1}{r}{\textbf{Total}} \\ \midrule
        \multirow{4}{*}{Train} & TyDiQA & 5,369 & 0 & 5,369 \\
         & Model Gen & 5,369 & 5,353 & 10,722 \\
         & Human Filt & 4,865 & 2,730 & 7,595 \\
         & IDK-MRC & 5,042 & 4,290 & 9,332 \\ \midrule
        \multirow{4}{*}{Dev} & TyDiQA & 402 & 0 & 402 \\
         & Model Gen & 402 & 401 & 803 \\
         & Human Filt & 364 & 211 & 575 \\
         & IDK-MRC & 382 & 382 & 764 \\ \midrule
        \multirow{4}{*}{Test} & TyDiQA & 423 & 0 & 423 \\
         & Model Gen & 423 & 423 & 846 \\
         & Human Filt & 405 & 249 & 654 \\
         & IDK-MRC & 422 & 422 & 844 \\ \bottomrule
    \end{tabular}%
    }
    \caption{Statistics of the datasets.}
    \label{tab:data-stat}
\end{table}

\section{Experiments}
We evaluate our IDK-MRC dataset by (1) comparing our automatic QG pipeline with several baselines to measure the performance of our automatic generation method, (2) analyzing the quality and cost of the automatic and manual dataset generation to see whether we can benefit from the additional manual/human-labeled data, and (3) comparing MRC models trained with IDK-MRC and others to validate the effectiveness of our dataset in the downstream task.

\subsection{Automatic Generation Model Evaluation}
\label{sec:qgen-experiment}
We compare our QG model to these methods:
\begin{itemize}
    \item \textbf{TF-IDF}: given an answerable question as the query, unanswerable question is generated by retrieving the most relevant question using TF-IDF features \cite{scikit-learn}. The similarity between the questions are calculated using cosine similarity.
    \item \textbf{Rule-based}: we replace the entity in the answerable question with another entity in the context that has the same type, i.e., an entity with type \texttt{PERSON} will be replaced by another entity with type \texttt{PERSON}. If there is no appropriate entity in the question, we randomly swap the question tag with another tag. We extract the entity using XLM-R\textsubscript{BASE} model\footnote{\url{https://huggingface.co/cahya/xlm-roberta-base-indonesian-NER}} and extract the question tag using a simple matching with our predefined question tag list.
    \item \textbf{Pair2Seq}: we adapt the pair-to-sequence model by \citet{zhu-etal-2019-learning} with Indonesian FastText \cite{bojanowski-etal-2017-enriching}. We follow the model architecture and the training procedure described in their paper. 
\end{itemize}

\paragraph{Dataset}
We use SQuAD 2.0 to train our QG model. First, we align answerable and unanswerable questions with the same plausible answer. For example, if there exists an answerable question such as:

\begin{quote}
    \textit{Who ruled the duchy of Normandy?}\newline
    Answer: \textit{\textbf{Richard}}
\end{quote}

\noindent
and an unanswerable question such as:

\begin{quote}
    \textit{Who ruled the country of Normandy?}\newline
    Answer: \texttt{[empty]}\newline
    Plausible answer: \textit{\textbf{Richard}}
\end{quote}

\noindent
the above questions will be paired or aligned.

Then, we translate the dataset using Google Translate API v2. Because complex questions tend to have more translation artifacts, we eliminate such questions by removing questions with a conjunction. From this process, we get 14,029 input pairs as the training data and 2,144 as the validation data. We use this dataset to train the question generation model (\S\ref{sec:automatic-gen}).

\paragraph{Implementation}
We implement QG model using SimpleTransformers \cite{rajapakse2019simpletransformers}. We use mT5\textsubscript{BASE} (580M parameters) to generate the questions with the maximum sequence length of 512. We train the model in 5 epochs and a batch size of 8. For the decoding, we use top-k and top-p sampling with a value of 50 and 0.95, respectively. We set the returned sequence number to 10.
\nocite{wolf-etal-2020-transformers}

\paragraph{Evaluation Metric}
We evaluate the models using the existing \textbf{BLEU} score metric\footnote{We use the NLTK version of the BLEU score (\url{https://www.nltk.org/api/nltk.translate.bleu_score.html}). The tokenization is done using the Indonesian tokenizer of Stanza library (\url{https://stanfordnlp.github.io/stanza}).}. However, BLEU is an n-gram based metric, so it can give a high score to the unanswerable questions that are exactly the same as the answerable question. Therefore, we use \textbf{\%diff} to compute the proportion of generated unanswerable questions that is not identical to its corresponding answerable questions. We also propose a new metric called \textbf{Unanswerable BLEU (UBLEU)}, an improved version of BLEU by setting the modified precision ($p_n$) to 0 if the output from the QG model ($q_{out}$) is identical to the paired answerable question ($q_{ans}$), formally defined as:
\[
p_{n} = \frac{
            \sum\limits_{C \in \{\textit{Candidates}\}} \sum\limits_{\textit{n-gram} \in C} \alpha \  \textit{Count}_{\textit{clip}}\left( \textit{n-gram} \right)
        }{
            \sum\limits_{C' \in \{\textit{Candidates}\}} \sum\limits_{\textit{n-gram}' \in C'} \textit{Count}\left( \textit{n-gram}' \right)
        }
\]
\noindent where
\[
\alpha = \left\{ 
  \begin{array}{ c l }
    0        & \quad \textrm{if } q_{out} = q_{ans} \\
    1  & \quad \textrm{otherwise}
  \end{array}
\right.
\]

\begin{table}[t!]
    \centering
    \resizebox{\columnwidth}{!}{%
    \begin{tabular}{@{}lrrrrr@{}}
        \toprule
        \multicolumn{1}{c}{\multirow{2}{*}{\textbf{Model}}} & \multicolumn{2}{c}{\textbf{UBLEU}} & \multicolumn{2}{c}{\textbf{BLEU}} & \multicolumn{1}{c}{\multirow{2}{*}{\textbf{\% diff}}} \\ 
        \multicolumn{1}{c}{} & \multicolumn{1}{c}{\textbf{3}} & \multicolumn{1}{c}{\textbf{4}} & \multicolumn{1}{c}{\textbf{3}} & \multicolumn{1}{c}{\textbf{4}} & \multicolumn{1}{c}{} \\ \midrule
        TF-IDF & 12.30 & 7.09 & 12.30 & 7.09 & \textbf{100\%} \\
        Rule-based & 41.26 & 33.92 & 41.26 & 33.92 & \textbf{100\%} \\
        Pair2Seq & 26.21 & 19.43 & 28.58 & 21.42 & 94.68\% \\ \midrule
        Ours & \textbf{43.81} & \textbf{36.50} & \textbf{43.97} & \textbf{36.63} & \underline{99.61\%} \\
        + QA Filter & 42.97 & 35.58 & 43.14 & 35.72 & 99.58\% \\ \bottomrule
    \end{tabular}%
    }
    \caption{Automatic evaluation of various QG models tested on translated version of SQuAD 2.0 dev dataset. \textbf{\%diff}: proportion of generated UnAns questions that is not identical (different) to the paired Ans questions; \textbf{BLEU}: BLEU score; \textbf{UBLEU}: Unanswerable BLEU.}
    \label{tab:result-qg-auto}
\end{table}
\begin{table}[t!]
    \centering
    \resizebox{\columnwidth}{!}{%
    \begin{tabular}{@{}lrrrrr@{}}
        \toprule
        \multicolumn{1}{c}{\textbf{Model}} & \multicolumn{1}{c}{\textbf{UnAns}} & \multicolumn{1}{c}{\textbf{Rel}} & \multicolumn{1}{c}{\textbf{Flue}} & \multicolumn{1}{c}{\textbf{Avg}} & \multicolumn{1}{c}{\textbf{\% prf}} \\ \midrule
        TF-IDF & 0.74 & 2.18 & \textbf{2.60} & \cellcolor[HTML]{EFEFEF}0.778 & \cellcolor[HTML]{EFEFEF}20\% \\
        Rule-based & 0.67 & \textbf{2.95} & 2.48 & \cellcolor[HTML]{EFEFEF}0.827 & \cellcolor[HTML]{EFEFEF}40\% \\
        Pair2Seq & 0.84 & 2.25 & 1.91 & \cellcolor[HTML]{EFEFEF}0.742 & \cellcolor[HTML]{EFEFEF}11\% \\ \midrule
        Ours & 0.79 & 2.90 & 2.45 & \cellcolor[HTML]{EFEFEF}0.858 & \cellcolor[HTML]{EFEFEF}50\% \\
        + QA Filter & \textbf{0.89} & 2.92 & 2.48 & \cellcolor[HTML]{EFEFEF}\textbf{0.897} & \cellcolor[HTML]{EFEFEF}\textbf{59\%} \\ \bottomrule
    \end{tabular}%
    }
    \caption{Human evaluation result from 100 randomly sampled unanswerable questions. \textbf{UnAns}: Unanswerability; \textbf{Rel}: Relevency; \textbf{Flue}: Fluency; \textbf{Avg}: Average of UnAns, Rel, Flue, normalized in 0--1 scale; \textbf{\% prf}: \% of samples with perfect UnAns, Rel, and Flue scores.}
    \label{tab:result-qg-human}
\end{table}

Moreover, we conduct human evaluation to further study the performance of the models. We randomly sample 100 questions for each QG models, and ask four annotators to evaluate the questions quality using the same protocol as \S\ref{sec:validation}.

\begin{figure*}[t!]
    \begin{center}
        \centerline{\includegraphics[width=\linewidth]{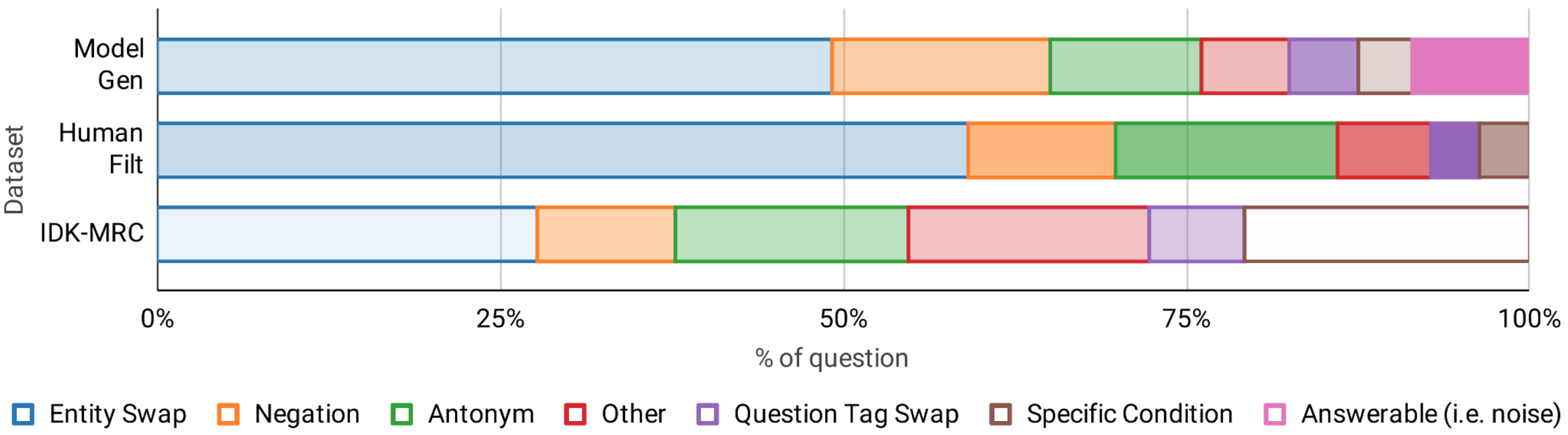}}
        \caption{Unanswerable question types distribution of Model Gen, Human Filt, and IDK-MRC test set. The question types are manually labeled by annotators. The bar opacity represents failure rate of the MRC model (XLM-R) in predicting the answer to the questions in each unanswerable question type (lighter is better). Our IDK-MRC dataset has a more balanced question type distribution, resulting in lower failure rate compared to Model Gen and Human Filt dataset.}
        \label{fig:test_distribution}
    \end{center}
\end{figure*}

\paragraph{Result}
As presented in Table \ref{tab:result-qg-auto} and \ref{tab:result-qg-human}, our QG model shows the best performance in both automatic and human evaluation. Despite a lower \%diff score than TF-IDF and rule-based, our model still achieves better UBLEU and BLEU scores. We also observe a slight reduction of UBLEU and BLEU scores when we add QA filter; however, based on human evaluation, QA Filter can improve the overall quality of the generated questions, especially the percentage of questions with perfect scores.

TF-IDF has a high fluency score because we use the existing answerable questions from different paragraphs, but it results in a low relevancy score. For rule-based, changing the entity in the answerable question to another entity in the context can produce high relevancy. However, it can still generate an answerable question, as shown by a lower unanswerability score. Pair2Seq \cite{zhu-etal-2019-learning} obtains a high unanswerability score but lower relevancy and fluency scores. It suffers from many \texttt{UNK} tokens, displaying the limitation of word embedding representation. Overall, adding QA Filter results in better performance in all evaluation aspects, indicated by a high average score and the number of samples with a perfect score.

\subsection{Automatic vs. Manual Generation}
We now compare the automatic and manual dataset generation from three perspectives: time, cost, and question quality, especially to further analyze whether the automatic generation model can benefit from additional human annotation.

\paragraph{Time and Cost}
For the automatic generation, it takes around 3 hours to train QG model on a single RTX 8000 48GB GPU. After the training has finished, the model takes 30 minutes to generate $\sim$2,000 questions in the inference step. As for the manual process, one person spent 32 hours verifying $\sim$2,000 questions and 10 hours writing $\sim$500 questions (40 hours per 2,000 questions). The cost for one human annotator is about \$7.5/hour, and assuming a GPU price of \$3/hour\footnote{The highest GPU hourly price from \url{https://cloud.google.com/compute/gpus-pricing} (Accessed June 2022).}, automatic generation is certainly more time- and cost-efficient approach than manual generation.

\begin{table*}[!t]
\centering
\resizebox{0.75\textwidth}{!}{%
\begin{tabular}{@{}clrrrr|r@{}}
\toprule
\multirow{2}{*}{\textbf{Model}} &
  \multicolumn{1}{c}{\multirow{2}{*}{\textbf{Train Dataset}}} &
  \multicolumn{2}{c}{\textbf{UnAns}} &
  \multicolumn{2}{c|}{\textbf{Overall}} &
  \multicolumn{1}{c}{\textbf{\begin{tabular}[c]{@{}c@{}}Avg UnAns\\ Failure Rate\end{tabular}}} \\
 &
  \multicolumn{1}{c}{} &
  \multicolumn{1}{c}{\textbf{EM}} &
  \multicolumn{1}{c}{\textbf{F1}} &
  \multicolumn{1}{c}{\textbf{EM}} &
  \multicolumn{1}{c|}{\textbf{F1}} &
  \multicolumn{1}{c}{\textbf{\%}} \\ \midrule
\multirow{5}{*}{IndoBERT} &
  Translated SQuAD &
  61.00 &
  61.00	& 
  52.42	&
  59.40 &
  45.96 \\
 &
  TyDiQA &
  0.19 &
  0.19 &
  31.00	&
  37.08 &
  98.47 \\
 &
  Model Gen &
  67.44 &
  67.44 &
  62.09 &
  67.45 &
  42.42 \\
 &
  Human Filt &
  66.64	&
  66.64	&
  62.49 &
  68.19 &
  52.35 \\
 &
  IDK-MRC &
  \textbf{86.26} &
  \textbf{86.26} &
  \textbf{72.06} &
  \textbf{77.45} &
  \textbf{31.92} \\ \midrule
\multirow{5}{*}{m-BERT} &
  Translated SQuAD &
  66.49 &
  66.49 &
  59.19 &
  65.48 &
  26.52 \\
 &
  TyDiQA &
  0.57 &
  0.57 &
  36.35 &
  41.23 &
  99.26 \\
 &
  Model Gen &
  79.10 &
  79.10 &
  72.35 &
  77.00 &
  19.16 \\
 &
  Human Filt &
  69.81 &
  69.81 &
  68.84 &
  73.73 &
  38.97 \\
 &
  IDK-MRC &
  \textbf{87.82} &
  \textbf{87.82} &
  \textbf{77.20} &
  \textbf{82.23} &
  \textbf{13.72} \\ \midrule
\multirow{5}{*}{XLM-R} &
  Translated SQuAD &
  66.78 &
  66.78 &
  59.00 &
  65.89 &
  26.13 \\
 &
  TyDiQA &
  0.90 &
  0.90 &
  33.01 &
  39.74 &
  97.32 \\
 &
  Model Gen &
  75.45 &
  75.45 &
  68.46 &
  74.40 &
  26.07 \\
 &
  Human Filt &
  67.87 &
  67.87 &
  64.95 &
  71.32 &
  41.96 \\
 &
  IDK-MRC &
  \textbf{88.29} &
  \textbf{88.29} &
  \textbf{74.86} &
  \textbf{81.37} &
  \textbf{16.42} \\ \bottomrule
\end{tabular}%
}
\caption{MRC models performance trained on various dataset. The EM and F1 scores are the models' performance on IDK-MRC test set, while the Avg Unanswerability Failure Rate are the models' performance on synthetic test cases generated using CheckList tool \cite{ribeiro-etal-2020-beyond}. We report average scores over 5 runs. The performance difference between the models trained on our dataset and the baselines are statistically significant ($p<0.05$).}
\label{tab:result-qa}
\end{table*}

\paragraph{Question Quality}
From Figure \ref{fig:test_distribution}, we observe that our automatic QG model manages to generate various unanswerable question types, as shown in Model Gen question types distribution. However, it still produces some noise, i.e., answerable questions (8.51\% of Model Gen test set), even after such questions are discarded by QA Filter model. We also observe that the QG model tends to produce more \textit{entity swap} questions (49.17\% of Model Gen test set). Moreover, many irrelevant or incomprehensible questions are still exist in Model Gen dataset, especially for \textit{negation}, \textit{question tag swap}, and \textit{specific condition} types. This result suggests that even though automatic QG model can generate relatively fluent and valid questions, relying \textit{only} on it for building the dataset may result in noise and imbalance question types distribution. Filtering out the noisy data, automatically or manually, is not enough since the question types distribution is still imbalanced, as can be seen in Human Filt question distribution. The additional human-written questions in IDK-MRC cover this limitation, resulting in a more balanced question type distribution and lower models' failure rate in predicting the answer for each unanswerable question type.

\subsection{Dataset Evaluation on Downstream Task}
Next, we investigate the performance of MRC models trained on our dataset and compare them with Translated SQuAD 2.0 \cite{muis2020seq2seq} and TyDiQA \cite{clark-etal-2020-tydi} datasets.

\paragraph{Implementation}
We pick IndoBERT\textsubscript{BASE} \cite{wilie-etal-2020-indonlu} as the monolingual model and m-BERT\textsubscript{BASE} \cite{devlin-etal-2019-bert}, XLM-R\textsubscript{BASE} \cite{conneau-etal-2020-unsupervised} as the multilingual model. They have 124.5M, 167.4M, and 278.7M parameters, respectively. We implemented the models using SimpleTransformers \cite{rajapakse2019simpletransformers}. We use the standard hyperparameter settings for QA task with maximum sequence length of 384, document stride of 128, and trained the models for 10 epochs, batch size of 8, learning rate of 2e-5 using the Adam optimizer \cite{kingma2014adam}.

\paragraph{Result on IDK-MRC test set}
We tested several models using IDK-MRC test set\footnote{We use IDK-MRC test set since there is no suitable existing dataset to show the performance on Indonesian unanswerable questions. Another option is to test the models on Translated SQuAD 2.0 test set, but we found that many questions have machine translation error or incomplete or wrong ground truth answer (43\% out of 100 randomly sampled questions). Therefore, Translated SQuAD 2.0 is not an adequate dataset to test Indonesian MRC models.} and the result is presented in Table \ref{tab:result-qa}. We observe that the models trained on IDK-MRC dataset perform better than all baselines. Furthermore, even the models trained on our less-cleaned Model Gen dataset can obtain a better result than Translated SQuAD dataset, indicating the importance of a dataset that originates in Indonesian. We also note that the models trained on TyDiQA fail to handle unanswerable questions, as shown by extremely low UnAns scores. This result further highlights the significance of incorporating unanswerable questions in MRC dataset.

\paragraph{Unanswerable failure rate}
To further examine the models' capability in handling unanswerable questions, we also conduct an unanswerability error analysis using CheckList \cite{ribeiro-etal-2020-beyond}, a tool that facilitates behavioral test on many NLP tasks. CheckList provides a list of linguistic capabilities, with each capabilities are divided into different \textit{test types} to further break down the potential \textit{failure} of the linguistic capabilities. In this experiment, we focus on testing the models' capability on predicting the answer to unanswerable questions by dividing the \textit{test types} into the unanswerable questions type listed in Table \ref{tab:data-sample}. The test cases for each test type are automatically generated using CheckList's \texttt{template} function, resulting in 600 test cases for each question type. The template examples are presented in Appendix \ref{appendix:checklist_qtype}.

As shown in last column of Table \ref{tab:result-qa}, we find that our dataset can reduce the average failure rate, further confirming the effectiveness of IDK-MRC compared to the existing dataset. IndoBERT has the most significant failure rate reduction (45.96 to 31.92), followed by m-BERT (26.52 to 13.72) and XLM-R (26.13 to 16.42). Also, it is clear that only relying on the existing TyDiQA dataset is not enough to build a robust model, as shown from very high failure rates.

Overall, most failures occur on \textit{negation} questions, specifically when the negation word appears in the context passage, such as:

\begin{quote}
    Context: \textit{Wikia \textbf{tidak} diketuai oleh Ali. (Wikia \textbf{is not} chaired by Ali.)}\newline
    Question: \textit{Siapa yang mengetuai Wikia? (Who is Wikia's chair?)}\newline
    Predicted Answer: \textit{Ali}\newline
    Correct Answer: \texttt{[empty]}
\end{quote}

Besides adding more data samples, we conjecture that some improvement in model architecture or training scheme is needed to solve this problem. It is possible that the model highly correlates "\texttt{who}" question tag with any person's name that appears in the context, and picks it as the answer without considering the meaning of the whole context. Additionally, the high failure rates on IndoBERT model are mainly contributed by the \textit{antonym} and \textit{question tag swap} types, while multilingual models like m-BERT and XLM-R performs significantly better on this type of question. All in all, focusing on better approach to handle the aforementioned question types for future work may further improve the models' performance.

\section{Conclusion}
We have presented IDK-MRC, the first Indonesian MRC dataset covering answerable and unanswerable questions. We confirm the effectiveness of our dataset in improving the MRC models' capability to handle unanswerable questions compared to other existing MRC datasets, such as Translated SQuAD and TyDiQA. We also verify that our automatic dataset generation method can help reduce the time and cost of the dataset collection. Subsequently, human supervision helps eliminate the dataset noise and question type imbalance problem from the automatic generation method. 

Although our dataset collection pipeline is designed to build unanswerable questions for Indonesian, it can also be utilized for other medium- to low-resource languages or other QA tasks, such as adversarial question generation. While our dataset pipeline (i.e., automatic generation, validation, manual generation) is general enough to be applied to other languages or QA tasks, further adjustment of the automatic question generation model is required. Still, we believe that our proposed pipeline has some potential to be generalized to several QA tasks, which may be an interesting direction for future work.

\section*{Limitations}
There may be some possible limitations in our study. Firstly, our automatic question generation (QG) model requires training data consisting of context paragraphs and answerable questions. Unlike medium- to low-resource languages like Indonesian, our QG method might be more challenging to be applied to extremely low-resource languages with even more limited data and resources.

Secondly, we utilized the existing transformer-based models that specifically pre-trained on Indonesian language, i.e., IndoBERT \cite{wilie-etal-2020-indonlu}. While we also used multilingual models like mT5 \cite{xue-etal-2021-mt5}, m-BERT \cite{devlin-etal-2019-bert}, and XLM-R \cite{conneau-etal-2020-unsupervised}, the number of languages covered by these models is also limited. Before applying those models to other language besides Indonesian, one must check whether the desired language exists during the pre-training phase of the models. Note that the model also needs to have high enough quality. Some of the large multilingual models are not very good for low- to extremely low-resource languages.

\section*{Ethics Statement}
The paragraphs and answerable questions that we utilized to build IDK-MRC dataset are taken from Indonesian subset of TyDiQA-GoldP dataset \cite{clark-etal-2020-tydi}, which originates from Wikipedia articles. Since those articles are written from a neutral point of view, the risk of harmful content is minimal. Also, all model-generated questions in our dataset have been validated by human annotators to eliminate the risk of harmful questions. During the manual question generation process, the annotators are also encouraged to avoid producing possibly offensive questions.

Even so, we argue that further assessment is needed before using our dataset and models in real-world applications. This measurement is especially required for the pre-trained language models used in our experiments, namely mT5 \cite{xue-etal-2021-mt5}, IndoBERT \cite{wilie-etal-2020-indonlu}, m-BERT \cite{devlin-etal-2019-bert}, and XLM-R \cite{conneau-etal-2020-unsupervised}. These language models are mostly pre-trained on the common-crawl dataset, which may contain harmful biases or stereotypes.

All datasets in this work are publicly available and distributed under CC BY-SA 4.0 license. Our data collection pipeline, along with the recruitment process of the human annotators, has been reviewed and approved by KAIST Institutional Review Board (KH2021-194). We ensured that annotators were paid above the minimum wage in the Republic of Korea.

\section*{Acknowledgements}
This work was partly supported by Institute of Information \& communications Technology Planning \& Evaluation (IITP) grant funded by the Korea government(MSIT) (No. 2022-0-00184, Development and Study of AI Technologies to Inexpensively Conform to Evolving Policy on Ethics). We also would like to thank Dea Adhista for managing the annotators during the validation and manual data collection process. Rifki Afina Putri was supported by Hyundai Motor Chung Mong-Koo Global Scholarship.

\bibliography{anthology,custom}
\bibliographystyle{acl_natbib}

\clearpage
\appendix
\renewcommand{\appendixpagename}{\Large \centering Appendix for "IDK-MRC: Unanswerable Questions for Indonesian Machine Reading Comprehension"}
\appendixpage
\section{Data Statement}

\subsection{Curation Rationale}
IDK-MRC dataset is built based on the existing paragraph and answerable questions (\texttt{ans}) in TyDiQA-GoldP \cite{clark-etal-2020-tydi}. The new unanswerable questions are automatically generated using the combination of mT5 \cite{xue-etal-2021-mt5} and XLM-R \cite{conneau-etal-2020-unsupervised} models, which are then manually verified by human annotators (\texttt{filtered ans} and \texttt{filtered unans}). We also asked the annotators to manually write additional unanswerable questions as described in \S\ref{sec:manual-gen} (\texttt{additional unans}). Each paragraphs in the final dataset will have a set of \texttt{filtered ans}, \texttt{filtered unans}, and \texttt{additional unans} questions. The illustration of the dataset collection pipeline is shown in Figure \ref{fig:data_collection}.

\subsection{Language Variety}
The texts in IDK-MRC are generated and written using the standard formal style of the Indonesian language.

\subsection{Annotator Demographic}
In our dataset collection pipeline, the annotators are asked to validate the generated unanswerable questions and write a new additional unanswerable questions.

We recruit four annotators with 2+ years of experience in Indonesian NLP annotation using direct recruitment. All of them are Indonesian native speakers who reside in Indonesia (Java Island) and fall under the 18--34 age category. We set the payment to around \$7.5 per hour. Given the annotators' demographic, we ensure that the payment is above the minimum wage rate (as of December 2021). All annotators also have signed the consent form and agreed to participate in this project.

\subsection{Speech Situation}
The paragraphs and answerable questions in IDK-MRC are built based on TyDiQA-GoldP \cite{clark-etal-2020-tydi}, which was originally taken from 2019 snapshots of Indonesian Wikipedia. As for the unanswerable questions, the dataset collection is conducted in December 2021. However, it is generated or written based on the facts or information provided in the existing paragraph in TyDiQA-GoldP dataset.

\subsection{Text Characteristics}
The original texts in IDK-MRC are mainly based from Wikipedia articles covering various topics, such as history, science, biography, and many more\footnote{The complete list of Indonesian Wikipedia article topics can be seen in \url{https://id.wikipedia.org/wiki/Wikipedia:Artikel_pilihan/Topik}}.

\subsection{Provenance Appendix}
As described in the previous section, the paragraphs and answerable questions in IDK-MRC are taken from the existing Indonesian TyDiQA dataset \cite{clark-etal-2020-tydi}. Unfortunately, the authors of TyDiQA did not provide complete data statement information, especially on their annotators demographic. However, since the original source of this dataset is from Wikipedia, we conjecture that the speech situation and text characteristic of this dataset is not far from the one that we have discussed in previous sections.

\section{Unanswerability Analysis by Question Type}
\label{appendix:checklist_qtype}
\begin{table}[t!]
    \centering
    \resizebox{\linewidth}{!}{%
    \begin{tabular}{@{}lp{5cm}@{}}
    \toprule
    \multicolumn{1}{c}{\textbf{Question Type}} & \multicolumn{1}{c}{\textbf{Template Example}}               \\ \midrule
    Negation (in question) & A is \texttt{VERB} by B. Who is not \texttt{VERB} B?                         \\
    Negation (in context)  & A is not \texttt{VERB} by B. Who is \texttt{VERB} B?                         \\
    Antonym                & A got the \texttt{ADJ} prize. Who got the \texttt{antonym of ADJ} prize? \\
    Entity Swap            & A is the president of B. Who is the president of C?      \\
    Question Tag Swap      & A was found on \texttt{DATE}. Who found A?                          \\
    Specific Condition                         & A is the president of B. Who is the (first) president of B? \\
    Other                  & A is \texttt{NOUN1} of B. Who is \texttt{NOUN2} of B?                  \\ \bottomrule
    \end{tabular}%
    }
    \caption{The test case template examples for '\texttt{who}' question tag.}
    \label{tab:checklist-template}
\end{table}
\begin{table*}[t!]
    \centering
    \resizebox{\textwidth}{!}{%
    \begin{tabular}{@{}lccccc|ccccc|cccccc@{\hskip3pt}l@{}}
        \toprule
        \multicolumn{1}{c}{\multirow{3}{*}{}} & \multicolumn{15}{c}{\textbf{Failure Rate}} & \multicolumn{2}{c}{\multirow{3}{*}{\textbf{\begin{tabular}[c]{@{}c@{}}Failure Cases Examples\\ with expected answer (A)\\ and model prediction (P)\end{tabular}}}} \\
        \multicolumn{1}{c}{} & \multicolumn{5}{c|}{\textbf{IndoBERT}} & \multicolumn{5}{c|}{\textbf{m-BERT}} & \multicolumn{5}{c}{\textbf{XLM-R}} & \multicolumn{2}{c}{} \\
        \multicolumn{1}{c}{} & \multicolumn{1}{c}{\textbf{SQ}} & \multicolumn{1}{c}{\textbf{TY}} & \multicolumn{1}{c}{\textbf{MG}} & \multicolumn{1}{c}{\textbf{HF}} & \multicolumn{1}{c|}{\textbf{IDK}} & \multicolumn{1}{c}{\textbf{SQ}} & \multicolumn{1}{c}{\textbf{TY}} & \multicolumn{1}{c}{\textbf{MG}} & \multicolumn{1}{c}{\textbf{HF}} & \multicolumn{1}{c|}{\textbf{IDK}} & \multicolumn{1}{c}{\textbf{SQ}} & \multicolumn{1}{c}{\textbf{TY}} & \multicolumn{1}{c}{\textbf{MG}} & \multicolumn{1}{c}{\textbf{HF}} & \multicolumn{1}{c}{\textbf{IDK}} & \multicolumn{2}{c}{} \\ \midrule
        \multirow{10}{*}{\rotatebox[origin=c]{90}{Negation}} & \multirow{5}{*}{0.3} & \multirow{5}{*}{99.4} & \multirow{5}{*}{\textbf{0.0}} & \multirow{5}{*}{0.4} & \multirow{5}{*}{0.9} & \multirow{5}{*}{0.1} & \multirow{5}{*}{99.6} & \multirow{5}{*}{\textbf{0.0}} & \multirow{5}{*}{\textbf{0.0}} & \multirow{5}{*}{1.1} & \multirow{5}{*}{\textbf{0.0}} & \multirow{5}{*}{95.2} & \multirow{5}{*}{\textbf{0.0}} & \multirow{5}{*}{\textbf{0.0}} & \multirow{5}{*}{\textbf{0.0}} & C: & Wikia dirancang oleh James. \\
         &  &  &  &  &  &  &  &  &  &  &  &  &  &  &  &  & \textit{Wikia was designed by James.} \\
         &  &  &  &  &  &  &  &  &  &  &  &  &  &  &  & Q: & Apa yg tidak dirancang James? \\
         &  &  &  &  &  &  &  &  &  &  &  &  &  &  &  &  & \textit{What was not designed by James?} \\
         &  &  &  &  &  &  &  &  &  &  &  &  &  &  &  & A: & \texttt{{[}empty{]}} P: Wikia \\ 
        \cmidrule(lr){2-18}
         & \multirow{5}{*}{\textbf{35.0}} & \multirow{5}{*}{100} & \multirow{5}{*}{66.8} & \multirow{5}{*}{83.5} & \multirow{5}{*}{60.9} & \multirow{5}{*}{\textbf{4.4}} & \multirow{5}{*}{99.5} & \multirow{5}{*}{55.5} & \multirow{5}{*}{84.1} & \multirow{5}{*}{19.4} & \multirow{5}{*}{\textbf{16.3}} & \multirow{5}{*}{94.6} & \multirow{5}{*}{53.9} & \multirow{5}{*}{77.8} & \multirow{5}{*}{49.6} & C: & Wikia tidak diketuai oleh Ali. \\
         &  &  &  &  &  &  &  &  &  &  &  &  &  &  &  &  & \textit{Wikia is not chaired by Ali.} \\
         &  &  &  &  &  &  &  &  &  &  &  &  &  &  &  & Q: & Siapa yg mengetuai Wikia? \\
         &  &  &  &  &  &  &  &  &  &  &  &  &  &  &  &  & \textit{Who is Wikia's chair?} \\
         &  &  &  &  &  &  &  &  &  &  &  &  &  &  &  & A: & \texttt{{[}empty{]}} P: Ali \\ 
        \midrule
        \multirow{5}{*}{\rotatebox[origin=c]{90}{Antonym}} & \multirow{5}{*}{\textbf{31.2}} & \multirow{5}{*}{100} & \multirow{5}{*}{63.9} & \multirow{5}{*}{88.7} & \multirow{5}{*}{74.4} & \multirow{5}{*}{\textbf{10.3}} & \multirow{5}{*}{99.9} & \multirow{5}{*}{20.7} & \multirow{5}{*}{45.6} & \multirow{5}{*}{16.8} & \multirow{5}{*}{\textbf{7.2}} & \multirow{5}{*}{99.3} & \multirow{5}{*}{24.9} & \multirow{5}{*}{41.0} & \multirow{5}{*}{22.5} & C: & Bia mendapatkan hadiah terendah. \\
         &  &  &  &  &  &  &  &  &  &  &  &  &  &  &  &  & \textit{Bia got the lowest prize.} \\
         &  &  &  &  &  &  &  &  &  &  &  &  &  &  &  & Q: & Siapa yg dapat hadiah tertinggi? \\
         &  &  &  &  &  &  &  &  &  &  &  &  &  &  &  &  & \textit{Who got the highest prize?} \\
         &  &  &  &  &  &  &  &  &  &  &  &  &  &  &  & A: & \texttt{{[}empty{]}} P: Bia \\ 
        \midrule
        \multirow{5}{*}{\rotatebox[origin=c]{90}{Ent Swap}} & \multirow{5}{*}{46.7} & \multirow{5}{*}{98.0} & \multirow{5}{*}{19.1} & \multirow{5}{*}{14.7} & \multirow{5}{*}{\textbf{14.2}} & \multirow{5}{*}{9.1} & \multirow{5}{*}{100} & \multirow{5}{*}{\textbf{0.8}} & \multirow{5}{*}{2.3} & \multirow{5}{*}{2.0} & \multirow{5}{*}{30.7} & \multirow{5}{*}{99.5} & \multirow{5}{*}{15.0} & \multirow{5}{*}{17.3} & \multirow{5}{*}{\textbf{5.9}} & C: & Dewi adalah presiden Kolombia. \\
         &  &  &  &  &  &  &  &  &  &  &  &  &  &  &  &  & \textit{Dewi is the president of Colombia.} \\
         &  &  &  &  &  &  &  &  &  &  &  &  &  &  &  & Q: & Siapa presiden Chili? \\
         &  &  &  &  &  &  &  &  &  &  &  &  &  &  &  &  & \textit{Who is the president of Chile?} \\
         &  &  &  &  &  &  &  &  &  &  &  &  &  &  &  & A: & \texttt{{[}empty{]}} P: Dewi \\ 
        \midrule
        \multirow{5}{*}{\rotatebox[origin=c]{90}{QTag Swap}} & \multirow{5}{*}{\textbf{48.8}} & \multirow{5}{*}{95.5} & \multirow{5}{*}{75.2} & \multirow{5}{*}{86.8} & \multirow{5}{*}{50.2} & \multirow{5}{*}{35.4} & \multirow{5}{*}{96.2} & \multirow{5}{*}{49.3} & \multirow{5}{*}{72.5} & \multirow{5}{*}{\textbf{27.4}} & \multirow{5}{*}{\textbf{20.5}} & \multirow{5}{*}{91.8} & \multirow{5}{*}{48.7} & \multirow{5}{*}{69.5} & \multirow{5}{*}{24.9} & C: & Anita lahir di Israel. \\
         &  &  &  &  &  &  &  &  &  &  &  &  &  &  &  &  & \textit{Anita was born in Israel.} \\
         &  &  &  &  &  &  &  &  &  &  &  &  &  &  &  & Q: & Kapan Anita lahir? \\
         &  &  &  &  &  &  &  &  &  &  &  &  &  &  &  &  & \textit{When was Anita born?} \\
         &  &  &  &  &  &  &  &  &  &  &  &  &  &  &  & A: & \texttt{{[}empty{]}} P: Israel \\ 
        \midrule
        \multirow{5}{*}{\rotatebox[origin=c]{90}{Specific Cond.}} & \multirow{5}{*}{74.6} & \multirow{5}{*}{99.9} & \multirow{5}{*}{26.4} & \multirow{5}{*}{38.0} & \multirow{5}{*}{\textbf{4.7}} & \multirow{5}{*}{75.0} & \multirow{5}{*}{100} & \multirow{5}{*}{6.5} & \multirow{5}{*}{37.7} & \multirow{5}{*}{\textbf{5.7}} & \multirow{5}{*}{62.8} & \multirow{5}{*}{100} & \multirow{5}{*}{18.2} & \multirow{5}{*}{40.8} & \multirow{5}{*}{\textbf{0.7}} & C: & Roy adalah seorang presiden. \\
         &  &  &  &  &  &  &  &  &  &  &  &  &  &  &  &  & \textit{Roy is a president.} \\
         &  &  &  &  &  &  &  &  &  &  &  &  &  &  &  & Q: & Siapa presiden paling terkenal? \\
         &  &  &  &  &  &  &  &  &  &  &  &  &  &  &  &  & \textit{Who is the most famous president?} \\
         &  &  &  &  &  &  &  &  &  &  &  &  &  &  &  & A: & \texttt{{[}empty{]}} P: Roy \\ 
        \midrule
        \multirow{5}{*}{\rotatebox[origin=c]{90}{Other}} & \multirow{5}{*}{56.9} & \multirow{5}{*}{97.7} & \multirow{5}{*}{36.6} & \multirow{5}{*}{44.0} & \multirow{5}{*}{\textbf{17.0}} & \multirow{5}{*}{27.1} & \multirow{5}{*}{99.9} & \multirow{5}{*}{\textbf{10.0}} & \multirow{5}{*}{33.6} & \multirow{5}{*}{20.3} & \multirow{5}{*}{27.5} & \multirow{5}{*}{98.5} & \multirow{5}{*}{22.7} & \multirow{5}{*}{44.3} & \multirow{5}{*}{\textbf{19.8}} & C: & Sheila adalah penggemar Rudy. \\
         &  &  &  &  &  &  &  &  &  &  &  &  &  &  &  &  & \textit{Sheila is Rudy's fan.} \\
         &  &  &  &  &  &  &  &  &  &  &  &  &  &  &  & Q: & Siapa teman Rudy? \\
         &  &  &  &  &  &  &  &  &  &  &  &  &  &  &  &  & \textit{Who is Rudy's friend?} \\
         &  &  &  &  &  &  &  &  &  &  &  &  &  &  &  & A: & \texttt{{[}empty{]}} P: Sheila \\ 
        \midrule
        \multicolumn{1}{c}{\textbf{}} & 45.96 & 98.47 & 42.42 & 52.35 & \textbf{31.92} & 26.52 & 99.26 & 19.16 & 38.97 & \textbf{13.72} & 26.13 & 97.32 & 26.07 & 41.96 & \textbf{16.42} & \multicolumn{1}{l}{} &  \\
        \bottomrule
    \end{tabular}%
    }
    \caption{The failure rate on all unanswerable question types tested using the CheckList tool \cite{ribeiro-etal-2020-beyond}. The scores are the mean over 5 runs with different random seeds (lower score is better). The \textbf{last row} denotes the \textbf{macro average} of all unanswerability types. \textbf{SQ}: translated SQuAD, \textbf{TY}: TyDiQA, \textbf{MG}: Model Gen, \textbf{HF}: Human Filt, \textbf{IDK}: IDK-MRC.}
    \label{tab:checklist}
\end{table*}
In this section, we aim to further measure the MRC models' performance on handling each unanswerable question type using CheckList tool.

\paragraph{Test Case Generation}
We utilized \texttt{template} function provided in Checklist to generate the test cases for each unanswerable question type, i.e., negation (in-question and in-context), antonym, entity swap, question tag swap, specific condition, and other. Each question type consists of several question tag, namely \textit{siapa} (who), \textit{apa} (what), \textit{kapan} (when), \textit{di mana} (where), \textit{mengapa} (why), and \textit{berapa} (how long/many/much). Each question tag have 100 test cases, therefore, we have a total of 600 test cases for each unanswerable question type. Some of the template examples are presented in Table \ref{tab:checklist-template}.

\paragraph{Experiment Result}
As shown from Table \ref{tab:checklist}, models trained on our IDK-MRC dataset has a lower failure rate on the \textit{entity swap}, \textit{question tag swap}, \textit{specific condition}, and \textit{other} questions, indicating that adding more examples for these question types can improve the models' unanswerability skills. Also, most models can successfully handle the negation if the negated word exists in the question. When the negated word appears in the context, most models fail to predict the correct answer. Meanwhile, models train on SQuAD has a lowest failure rate on negation case, and we conjecture that it occurs due to the imbalanced question type distribution in the SQuAD training dataset. As reported by \citet{sen-saffari-2020-models}, 85\% of questions containing "\texttt{n't}" and 89\% of questions containing "\texttt{never}" in SQuAD dataset are categorized as unanswerable question. It aligns with our experiment results, which shows that SQuAD has a lowest failure rate on negation question type and a much higher failure rate on the other question types.

\section{Annotation Instruction}
\label{appendix:instruction}

\begin{figure*}[t!]
    \begin{center}
        \centerline{\includegraphics[width=\linewidth]{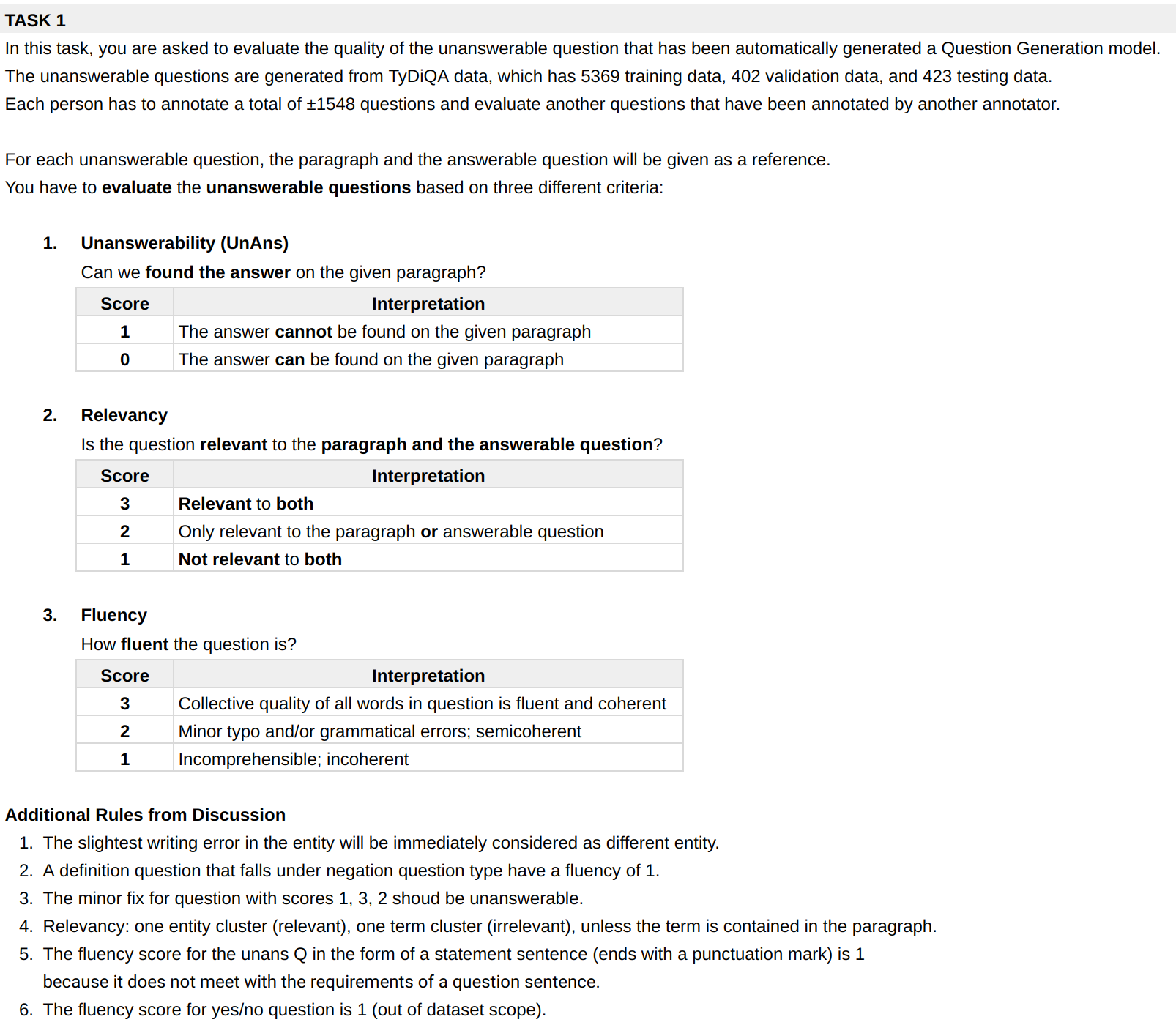}}
        \caption{Annotation instruction for the validation stage.}
        \label{fig:instruction_1}
    \end{center}
\end{figure*}

\begin{figure*}[t!]
    \begin{center}
        \centerline{\includegraphics[width=\linewidth]{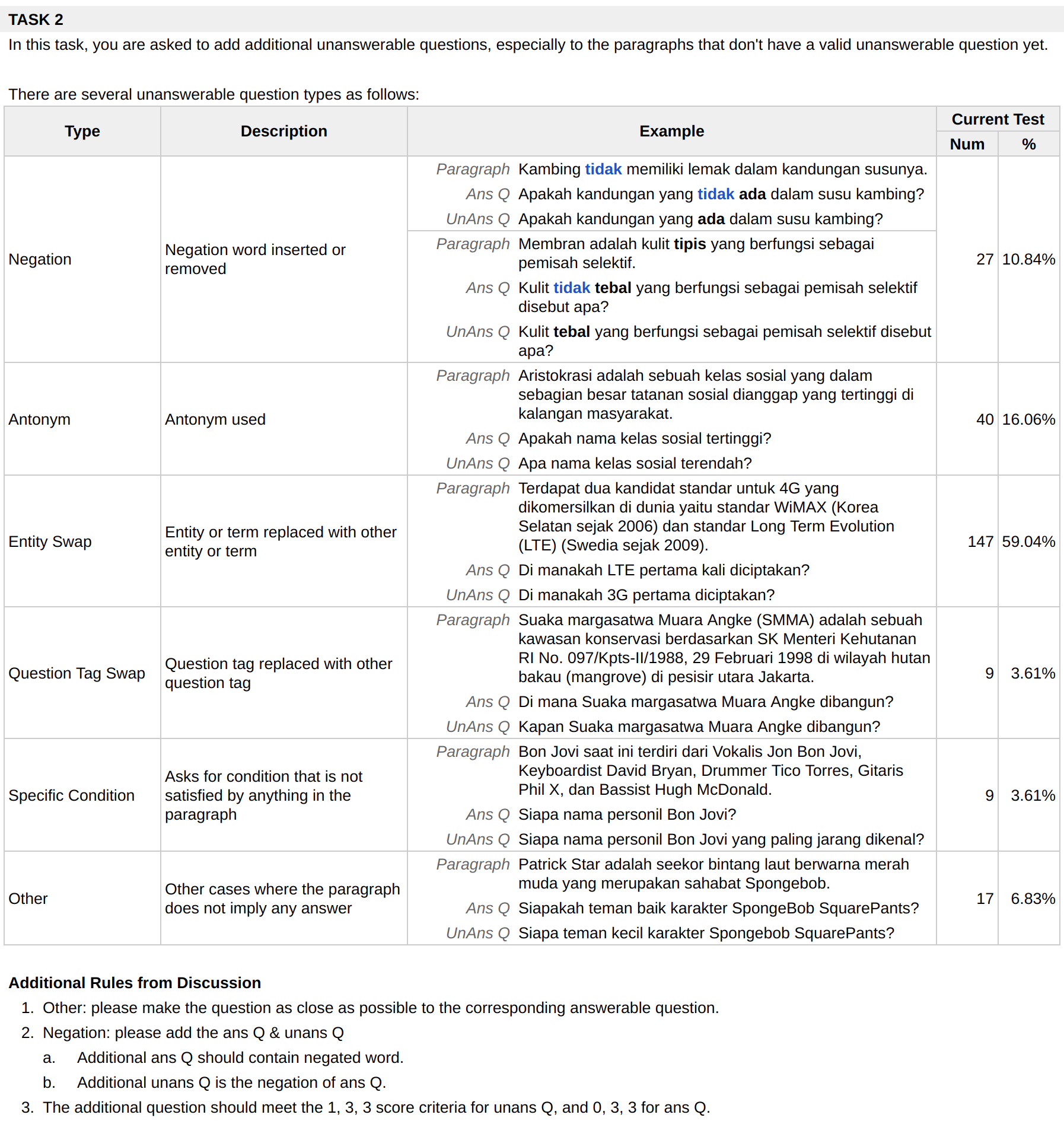}}
        \caption{Annotation instruction for the manual generation stage.}
        \label{fig:instruction_2}
    \end{center}
\end{figure*}

\subsection{Validation Stage}
In this stage, human annotators are asked to validate the quality of the model-generated unanswerable questions using criteria as described in \S\ref{sec:validation}. Detailed instruction can be seen in Figure \ref{fig:instruction_1}.

\subsection{Manual Generation Stage}
In this stage, human annotators are asked to write a questions that the question generation model fails to generate as described in \S\ref{sec:manual-gen}. The instruction given to the annotators are shown in Figure \ref{fig:instruction_2}.

\end{document}